\documentclass[journal]{IEEEtai}
\usepackage[colorlinks,urlcolor=blue,linkcolor=blue,citecolor=blue]{hyperref}

\usepackage{color,array}

\usepackage{graphicx}

\usepackage{multirow} 
\usepackage{algorithm}
\usepackage{algorithmic}

\begin{document}

\title{Automating Code Adaptation for MLOps - A Benchmarking Study on LLMs }

\author{Harsh Patel, Buvaneswari A. Ramanan,
Manzoor A. Khan,
Thomas Williams,
Brian Friedman, and
Lawrence Drabeck

\thanks{
    Harsh Patel was an intern with Nokia Bell Labs, Murray Hill, NJ 07974, USA. He is now with Dept of Computer Science and Engineering, University of California San Diego, La Jolla, CA 92092 USA (e-mail: h1patel@ucsd.edu).
    }
\thanks{
    Buvaneswari A. Ramanan, Manzoor A. Khan, Thomas Williams, Brian Friedman and Lawrence Drabeck are with Nokia Bell Labs, Murray Hill, NJ 07974, USA. (email: \{buvana.ramanan, manzoor.a.khan, thomas.williams, brian.friedman, lawrence.drabeck\}@nokia-bell-labs.com).
    }
}


\maketitle

\begin{abstract}

This paper explores the possibilities of the current generation of Large Language Models for incorporating Machine Learning Operations (MLOps) functionalities into ML training code bases. We evaluate the performance of OpenAI (gpt-3.5-turbo) and WizardCoder (open-source, 15B parameters) models on the automated accomplishment of various MLOps functionalities in different settings. We perform a benchmarking study that assesses the ability of these models to: 1) adapt existing code samples (\textit{Inlining}) with component-specific MLOps functionality such as MLflow and Weights \& Biases for experiment tracking, Optuna for hyperparameter optimization etc., and 2) perform the task of \textit{Translation} from one component of an MLOps functionality to another, e.g., translating existing GitPython library based version control code to Data Version Control library based. We also propose three different approaches that involve teaching LLMs to comprehend the API documentation of the components as a reference while accomplishing the \textit{Translation} tasks. 
In our evaluations, the gpt-3.5-turbo model significantly outperforms WizardCoder by achieving impressive Pass@3 accuracy in model optimization (55\% compared to 0\% by WizardCoder),  experiment tracking (100\%, compared to 62.5\% by WizardCoder), model registration (92\% compared to 42\% by WizardCoder) and hyperparameter optimization (83\% compared to 58\% by WizardCoder) on average, in their best possible settings, showcasing its superior code adaptability performance in complex MLOps tasks.
\end{abstract}

\begin{IEEEImpStatement}

MLOps is a set of processes and operations that enhances the accessibility of ML-based applications and empowers  organizations to embrace AI. MLOps-driven innovation integrates AI seamlessly into daily life and assures fairness and transparency. MLOps enhances efficiency, cost-effectiveness, and environmental sustainability.  However,  incorporating MLOps into the business use cases imposes substantial costs, both in terms of financial investments and resource allocation, as these practices require targeted training to the developers and  substantial time and effort for adaptation.  


The impact of our work may be summarized as: i) empowering enterprises and users to choose ML components based on their preferences, ii) lowering the barrier to the adoption of MLOps by facilitating its rapid deployment, accelerating AI adoption across businesses iii) empowering the system integrators to cater to diverse requirements of enterprises, iv) enhancing the product portfolio of equipment and solution providers by offering feature-enriched solutions without incurring high costs.

\end{IEEEImpStatement}


\section{Introduction}

\IEEEPARstart{D}{eploying} machine learning models in production has been a long challenge since the inception of the earliest models. Various machine learning efforts fail, with many proofs of concepts never making it to production \cite{VB}. The concept of Machine Learning Operations (MLOps) has emerged as a comprehensive approach to streamline and automate the deployment, monitoring, and maintenance of machine learning models in production. As the field of Machine Learning (ML) advances, MLOps tools and practices continue to evolve and minimize the challenges associated with deploying ML models in production. The impact of MLOps is evident through reduced deployment times \cite{gift}, improved model robustness \cite{sus}, and better collaboration between data scientists \cite{mlops}. 

However, with the advent of novel and better tools for several ML functionalities, quickly and effectively integrating them into existing MLOps services can be a challenge \cite{challenge}. In this work, we explore the feasibility of automatically accomplishing several MLOps functionalities such as model optimization, experiment tracking, model registration and hyperparameter optimization with use of Large Language Models (LLMs). Moreover, we also demonstrate the ability of LLMs in enabling ML Engineers / Data Scientists to adapt their existing MLOps pipelines with emerging new components.
    
Following the success of LLMs such as \cite{bert, gpt} on several natural language processing tasks such as text completion, question-answering, etc., there has been a substantial surge in research works on LLMs for code-related tasks. These tasks include code retrieval \cite{codesearch, codebert}, code generation \cite{progsynth, univmath}, code completion \cite{codefill} etc. HumanEval \cite{humaneval}, the most popular code synthesis evaluation framework, consists of 164 original programming problems that assess language comprehension, algorithms, and simple mathematics to simple software interview questions. Moreover, \cite{progsynth} conducted a large-scale study of how LLMs perform at synthesis of short Python programs. We identify a lack of evaluation studies that assess the ability of these  emerging code LLMs on complex instruction-based code adaptation tasks. Our work on assessing the ability of LLMs for automating MLOps is aimed at bridging this gap.

In particular, this paper makes the following contributions:
\begin{itemize}
    \item Our research is the first initiative towards the automated code adaptation of critical MLOps-related development tasks of model optimization, experiment tracking, model registration, data version control, and hyperparameter optimization. 
    \item We are the first research group to introduce the integration of LLMs as a fundamental tool to achieve this automation, emphasizing on its significant potential.  
    \item We introduce two distinct task categories that evaluate the capacity of LLMs, addressing their ability to adapt code to multiple MLOps functionalities and tackle complex code adaptation challenges. The first task category (\textit{Inlining}), involves adapting existing code examples by making changes within the code or "in-between-the-lines" using LLMs. The second task category (\textit{Translation}) involves translating existing code examples written using one component to another available component of an MLOps functionality.
    \item We perform a comprehensive benchmarking study that evaluates the performance of both open and closed-source LLMs in the automated adaptation of existing codebases to diverse MLOps functionalities. This study incorporates task-specific knowledge enhancements and evaluates the performance of LLMs at different temperatures.
    \item We propose three novel methodologies that enhance the API documentation comprehension within LLMs significantly, especially in the context of the \textit{Translation} task category, thereby exemplifying the practical application of LLMs in real-world scenarios.    
    \item We provide valuable insights and best practices, aimed at empowering practitioners to effectively utilize the capabilities of LLMs, whether working in the MLOps, DevOps, or Software Engineering domains.
\end{itemize}

\section{Background}


\begin{figure}[t]
\centering
\includegraphics[width=0.95\columnwidth]{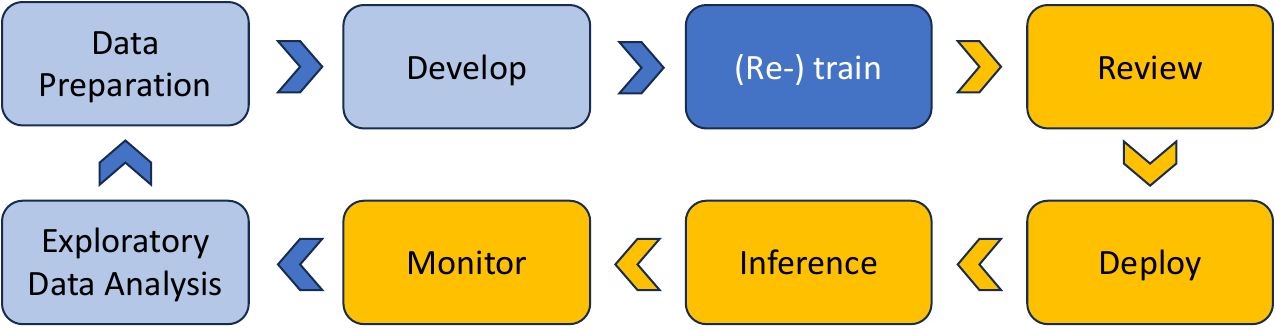} 
\caption{The iterative process of ML development}
\label{fig:MLOPs}
\end{figure}

An organization's ML pursuit often begins with a specific problem that requires data-driven solutions, and ML emerges as the appropriate tool. Initially, a significant amount of time goes into collecting, integrating, and thoroughly cleaning data from various sources. The next phase involves feature engineering, model training, and rigorous model evaluations. Data scientists experiment with different models, configurations, and architectures to optimize performance. If the models perform well, they move to the next step of deploying the model for generating predictions. However, the most crucial step revolves around continuous model monitoring. Real-world data dynamics necessitate continuous monitoring to ensure sustained model effectiveness. Monitoring helps identify when it's time to update the model or gather new data,  which leads to the next iteration of the ML model development and deployment. The iterative process of the ML model life-cycle is illustrated in Fig. \ref{fig:MLOPs}. As organizations increasingly adopt ML, manual workflows for building, training, and deploying models can become bottlenecks. Implementing effective operational practices is crucial for productivity, sustained model performance and robustness.

\begin{figure*}[t]
\centering
\includegraphics[width=\linewidth]{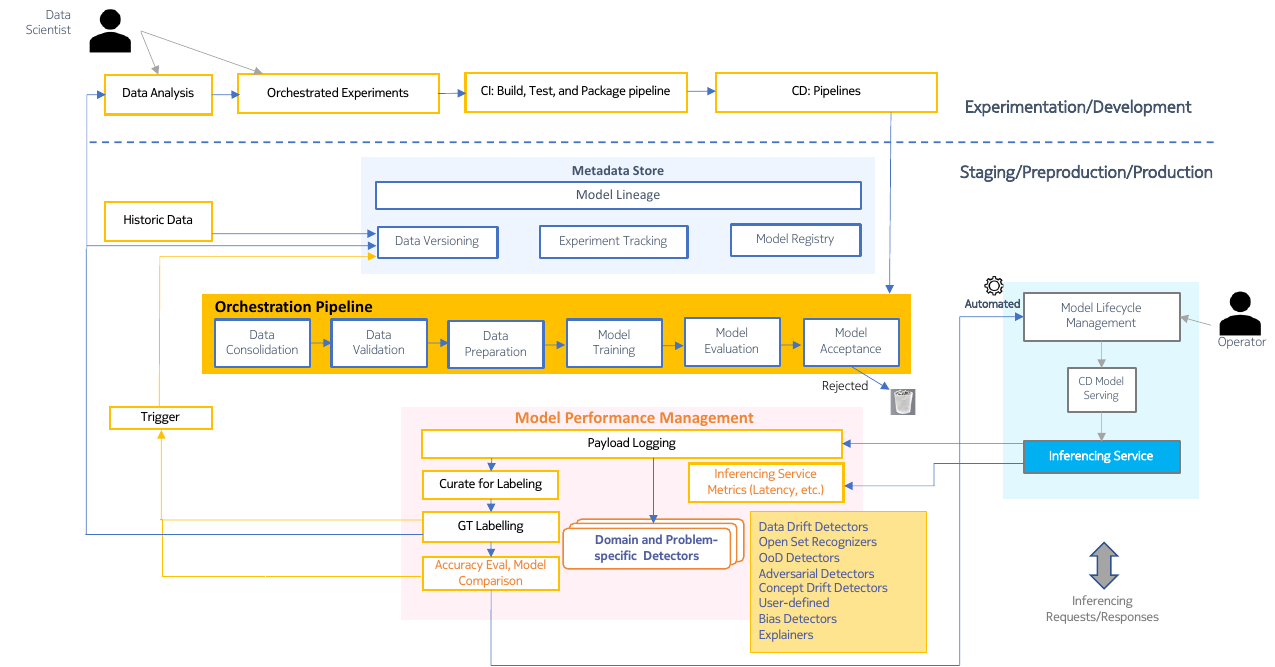} 
\caption{An end-to-end Machine Learning Operations (MLOps) System.}
\label{fig:mlaas}
\end{figure*}

MLOps is a set of operational practices to automate and standardize model building training, deployment, monitoring, management, and governance. It is at the intersection of the fields of DevOps, ML and Data Engineering. MLOps can help companies streamline the end-to-end ML life-cycle and boost productivity of data scientists and MLOps teams while maintaining high model accuracy and enhancing security and compliance. It involves the inter-working of four major subsystems and the constituent components thereof, as illustrated in {Fig. \ref{fig:mlaas}}. 

\subsection{Orchestrated pipeline for retraining}
The ML model is automatically trained on the dataset that is a result of a data consolidation and validation process. The training may involve Hyperparameter Optimization, and the trained models are evaluated on the test dataset and are sent for model acceptance testing. Accepted models are delivered to the inferencing services continuously, using Continuous Delivery (CD) practices such as GitOps. The following components are central to this subsystem:


\subsubsection{Data Processing / Transformation}
   Data Processing involves data cleaning, normalization, feature engineering, and data augmentation. It often includes handling missing values and outliers, as well as converting categorical data into numerical form. Popular tools and frameworks are Pandas, NumPy, Scikit-Learn, TensorFlow Data Validation (TFDV), Apache Spark, and Apache Flink.

\subsubsection{ML Model Training}
    ML Model Training is the core of machine learning development. It includes selecting algorithms, splitting data into training and validation sets, hyperparameter tuning, and training models. Deep learning models require powerful hardware like GPUs or TPUs. Popular frameworks/tools are Scikit-Learn, TensorFlow, PyTorch, Keras, XGBoost, LightGBM, and H2O.ai.

\subsubsection{Model Optimization / Compression (Pruning and Quantization)}
    Model Compression techniques aim to reduce the size of trained models to make them more efficient for deployment. Pruning removes unnecessary connections or layers, while quantization reduces the precision of model weights. Hardware-specific optimizations can further improve model efficiency. Popular frameworks/tools are TensorFlow Model Optimization Toolkit (TFLite), PyTorch quantization libraries, NNCF and Torch Pruning.

\subsection{Model Inferencing}

Inferencing is the stage when the trained ML model is used to make a prediction against unseen or in-the-wild data, and is usually accomplished by standing up the model for serving in an inferencing graph (IG). Lifecycle management refers to the lifecycle of the IG and is where decisions about the structure of the graph are made. For example, in shadow routing, which model will be treated as the primary, and which ones will be the shadows. Inferencing requires serving infrastructure and efficient model execution, often on specialized hardware and/or using optimized runtimes. Popular frameworks/tools are TensorFlow Serving, Seldon and BentoML.

\subsection{Model metadata management}

The purpose of this subsystem is to keep track of the essential metadata about the ML models, and the datasets to get the associations between a model, and all the components that were involved in the creation of that model, including container images, configurations, and the processes involved as well as in the comparison of the various models and the datasets. The following components are central to this subsystem:

\subsubsection{Experiment Tracking}
Experiment Tracking records metadata about model training runs, such as hyperparameters, metrics, and configurations. It helps in managing experiments, reproducing results, and comparing different models. Popular frameworks/tools are MLflow, Neptune, Comet and Weights \& Biases.

\subsubsection{Model Registry}
Model Registry manages versions of trained models, tracking model lineage and metadata. It ensures that models are properly documented and can be easily traced, deployed, or rolled back. Popular frameworks/tools are MLflow Model Registry, and ModelDB.

\subsubsection{Data Version Control}
    Data Version Control tracks changes to datasets, ensuring data reproducibility across the ML pipeline. It helps synchronize data with the model development process and enables collaboration on datasets. Popular frameworks/tools are DVC (Data Version Control), Git LFS and LakeFS

\subsection{Model Performance Management}

The objective is to facilitate model performance monitoring and model continual training. There are two parts to the performance monitoring – the first refers to the basic inferencing metrics such as latency and throughput. The other, more ML-specific part is empowered by the payload logging facility of the inferencing service. Advanced algorithms, which can also be ML-based, are employed to perform drift detection, out-of-distribution detection, adversarial attack detection, etc. The detection results are used as triggers for curating data for continual training as well as for invoking the ML-pipeline. The following components are central to this subsystem:

\subsubsection{Detectors for Model Monitoring}
    Deployed models are continuously monitored by sending a copy of the inference request and  response to a variety of detectors to detect concept drift, data drift, model performance changes, etc,. Popular libraries are Seldon Alibi, and Evidently.
\subsubsection{Ground Truth Labelling}
    Ground Truth Labelling involves obtaining high-quality labels for training datasets. It can be a manual or automated process, crucial for supervised learning. Annotators ensure that labeled data is accurate and consistent. Amazon Mechanical Turks, Label-studio and AnyLabeling are a few existing solutions for labelling.

\section{Our Approach for MLOps Automation}



Adapting an existing codebase to support the MLOps functionalities outlined in the previous section may require meticulous tweaking of the code structure, and can demand time-consuming, tedious, and repetitive coding tasks. Additionally, the type of adaptation for a given functionality can vary from component to component. Our research question is to find out the extent to which the code adaptation tasks could be automated. 

Our approach is to utilize LLMs in streamlining the process of integrating MLOps functionalities into existing ML code. With LLMs, developers can automate various repetitive tasks, including the integration of version control, experiment tracking, model registration, and model optimization capabilities. In addition, LLMs can also provide valuable insights and recommendations for optimizing the MLOps workflow, enabling developers to enhance efficiency and scalability in the deployment and management of ML models.

Let us explore the intricacies of the code adaptation process for each MLOps functionality, which can highlight the rationale behind advocating for the utilization of LLMs in this context.

\subsection{MLOps Task Categories}
At a high level, the adaptation tasks can be grouped into two main categories as follows:
\subsubsection{Inlining} 
\begin{figure*}[t]
\centering
\includegraphics[width=0.9\textwidth]{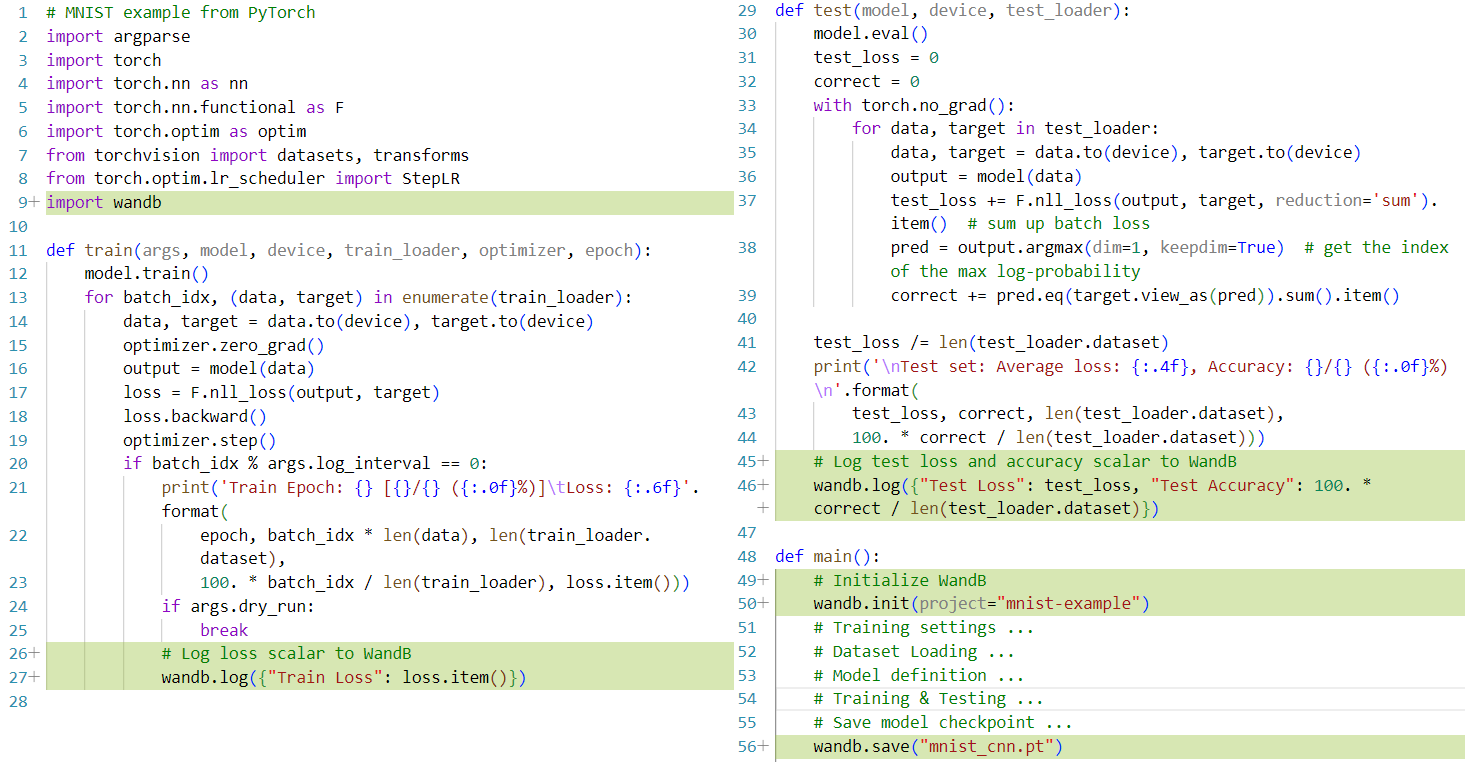} 
\caption{Code Inlining Task - Experiment Tracking. The highlighted green sections here demonstrate the expected inline adaptations when a model training script is provided as an input to an LLM for accomplishing experiment tracking using Weights \& Biases Library. Some parts of the code are skipped for better representation in the paper (shown as comments)}
\label{fig:exp-tracking}
\end{figure*}

\begin{figure*}[t]
\centering
\includegraphics[width=0.9\textwidth]{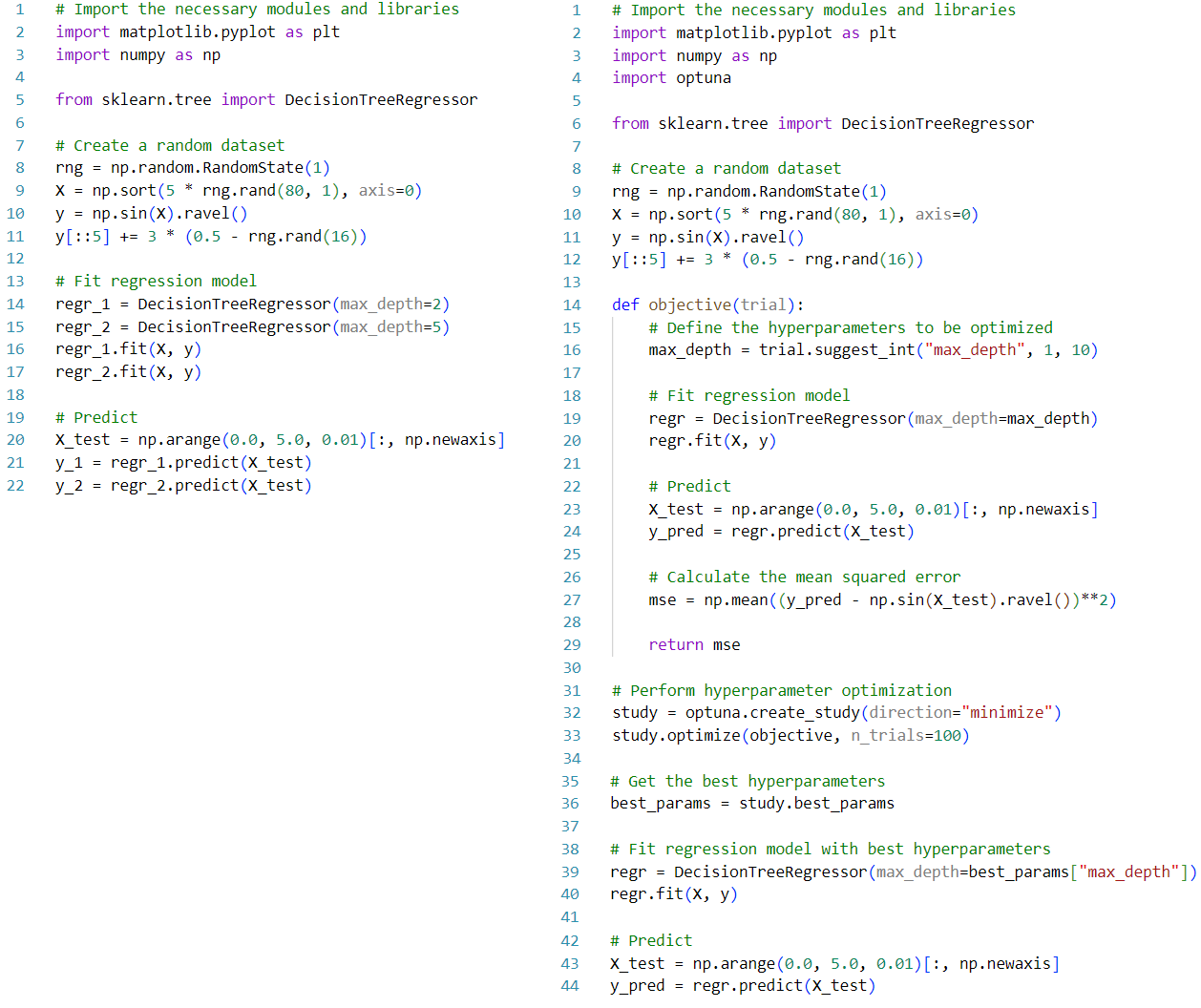} 
\caption{Code Inlining Task - Hyperparamter Optimization using Optuna Library. The left side of the figure displays a simple code snippet based on the scikit-learn library for performing Decision Tree Regression on random data. The right side demonstrates the expected output when utilizing an LLM to automatically accomplish hyperparameter optimization.}
\label{fig:hpo}
\end{figure*}

This category involves the adaptation of existing code by embedding task-specific changes. Within this category, we explore the following important MLOps functionalities:
        \begin{itemize}
            \item \textbf{Model Optimization}:
            Improving the efficiency of ML models using techniques such as pruning that reduce their size and computational complexity. Optimizing models can lead to improved accuracy, faster training times, and reduced resource consumption. Fig. \ref{fig:mo_example} demonstrates an example PyTorch-based training code adapted to incorporate model pruning using Neural Network Compression Framework (NNCF) \footnote{https://github.com/openvinotoolkit/nncf}.
            The adaptations include importing required modules from the nncf library, loading a configuration file that contains the desired NNCF settings (such as desired compression technique, pruning rate etc.). Further, `create\_compressed\_model` function is used to create a compressed model, along with an associated controller, that enables the application of advanced compression strategies while retaining the model's functionality. Furthermore, inside the training loop, the loss corresponding to the compression is integrated alongside the main training loss to ensure that the optimization process accounts for both, facilitating the effective training of the main model with integrated model pruning capabilities.

            \item \textbf{Experiment Tracking}: 
            This is the process of logging and monitoring experiments conducted during the model development cycle. Logging model configurations, hyperparameters, and metrics can aid in model reproducibility, comparison, and explainability. To better understand how to adapt a training code to support this functionality, Fig. \ref{fig:exp-tracking} demonstrates an example that uses the Weights \& Biases (wandb) library to accomplish the task. The adaptations include importing the \textit{wandb} library; initializing a new background process using \textit{wandb.init(project)} to log data to a run inside a given project. Further, Weights \& Biases functionalities such as \textit{wandb.log()}, \textit{wandb.save()}, \textit{wandb.watch()},  \textit{wandb.Image()} etc., are used for automatically logging relevant metrics, saving artifacts, watching gradients, system metrics and logging data samples corresponding to the run. This involves meticulously choosing the relevant variables to log as well as the position at which they should be logged such that it can provide a comprehensive view of the model's performance over the training time. The programming tasks for incorporating the experiment tracking using MLflow and Comet libraries are also along similar lines.  

            \item \textbf{Model Registration}:
            This is the process of managing and keeping track of different versions of ML models so that they can be easily retrieved and deployed. 
            To adapt an existing code for model registration using MLflow library, one must initialize a new MLflow run using \textit{mlflow.start\_run()} and train the model as per the existing code. At the end of training, the model can be logged in multiple programmatic ways. First, \textit{mlflow.\textless model\_flavor\textgreater.log\_model(model, registered\_model\_name)} method is used to register a model into a Model Registry for some predefined model flavors in the MLflow library such as Pytorch Lightning, Sklearn, Keras etc. In this method, if a registered model with the name does not exist, the method registers a new model and creates Version 1 into the Model Registry. If a registered model with the name exists, the method creates a new model version. The second way is to use the \textit{mlflow.register\_model(model\_uri)} method to create a new model version in model registry for the model files specified by \textit{model\_uri}. For this method, the \textit{run\_id} corresponding to this run is required to determine the \textit{model\_uri} referring to the model's directory. The procedure of code adaptation for other components such as Weight \& Biases is similar to that of MLflow.

            \item \textbf{Hyperparameter Optimization (HPO)}: This is the process of finding the best combination of hyperparameters (e.g., learning rate, batch size) to achieve optimal model performance. Effective hyperparameter tuning can significantly impact a model's accuracy and convergence speed.
            As for the actual process of incorporating HPO support into existing training code, several modifications need to be implemented to enable dynamic exploration of different parameter configurations. Fig. \ref{fig:hpo} illustrates an example of incorporating HPO into an existing scikit-learn based code using Optuna library, which is known for its efficient and versatile optimization algorithms.
            Within the training code, a new objective function is defined to encapsulate the model training process. This function dynamically adjusts the hyperparameters that need to be optimized (such as number of layers in a neural network based model, learning rate, maximum depth for decision trees and more) and return a score that can be maximised or minimised as desired. To facilitate this optimization, \textit{optuna.create\_study(direction)} method is used to create a study object. The study is then optimised using the \textit{optimize()} function which explores the trial suggestions provided by Optuna. It then evaluates different combinations of hyperparameters and determines the best set of hyperparameters to achieve the optimization goal. These optimized hyperparameters can be further used to train the final model, ultimately improving the performance for the given task.            
            
        \end{itemize}
 \subsection{Translation}  This task category involves the translation of existing code examples written using one component to another component of an MLOps functionality. Note that even though this may sound simple, the challenge comes when the LLM does not have knowledge about the target component at the time of its training. In addition, the absence of a direct correspondence between the APIs of the source and target components from a functional standpoint presents an additional challenge. In our work, we explore the following important MLOps functionalities:
    \begin{itemize}
            \item \textbf{Version Control}: Involves systematically managing changes made to all the project assets over time, including models, datasets and configuration files. Using a version control can immensely help to track changes, make updates or even roll back changes. This MLOps functionality can be achieved using components such as GitPython, DVC API,  LakeFS etc. The code adaptation task involves the rewriting of the API calls of one component to those of the other component.
    \end{itemize}
    

Mastering the code adaptation tasks necessitates a significant investment of time and effort, as developers must grasp the functionalities of various MLOps tools and comprehend their application within the context of specific use cases. Moreover, the potential for human errors during the implementation brings forth a need for a reliable and efficient solution that can streamline the adaptation process. In this context, the utilization of LLMs emerges as an alternative, given their capability to automate and expedite repetitive coding tasks, enabling developers to focus on higher-level conceptualization and problem-solving aspects of model development and optimization. As can be seen from the examples, the repetitive nature and standardized structure of code adaptations across various use cases make the employment of LLMs a promising approach.
\section{Methodology of LLM-based Automation}

Our research aims to derive a set of guidelines and best practices for effective and efficient MLOps automation using LLMs by utilizing carefully curated datasets and appropriate prompt engineering techniques tailored to the specific requirements of the tasks. The subsequent subsections will delve into the details of the essential elements, namely, the models, datasets, code adaptation (with the different prompt tuning techniques), and evaluation methodologies.

\subsection{Models} The models we use in this paper are dense decoder-only Transformer language models \cite{Vaswani2017AttentionIA} trained on a variety of data from the web. Open AI's gpt-3.5-turbo is a successor of their GPT-3 model which is trained on a filtered version of the CommonCrawl \cite{gpt3} data along with other Wikipedia and Books datasets, constituting almost three trillion tokens. We use the \textit{gpt-3.5-turbo-16k} model which has a context length of 16,000 tokens enabling experiments with large input examples.

WizardCoder is an instruction fine-tuned version of StarCoder \cite{starcoder} on a complex synthesized dataset. This data is generated using an EvolInstruct method \cite{wizardlm} that involves using LLMs instead of humans to automatically mass-produce open-domain instructions of various difficulty levels. The base StarCoder model is trained on permissively licensed data from GitHub that includes 80+ programming languages, Git commits, GitHub issues, and Jupyter notebooks. It is a ~15B parameter model that has a context length of over 8,000 tokens and is trained on around 1 trillion tokens.

\subsection{Datasets}
Since we are performing a benchmarking study to assess the performance of LLMs in MLOps-based code generation tasks, it is important to cover a diverse spectrum of code examples to show the unbiased performance of the models.  We focus on the following factors - popularity: use examples that are from popular frameworks and also in tutorials of popular libraries, - complexity: use of simple to complicated models, small to large length code, and reproducibility.

In conducting our benchmarking study to evaluate the performance of LLMs in MLOps-based code generation tasks, we recognize the importance of inclusivity and representativeness in our dataset selection. To ensure an unbiased and comprehensive assessment of model capabilities, we take into account the following critical factors:
\begin{itemize}
    \item We choose code examples that are based on popular frameworks such as PyTorch, Keras, sklearn, PyTorch Lightning, etc. Additionally, we consider examples that are featured in the tutorials offered by library developers. This emphasis on popularity reflects real-world usage scenarios and established best practices within the MLOps community.
    \item Our dataset encompasses a wide spectrum of complexity levels. We intentionally include examples that range from relatively simple, straightforward models such as Basic Convnets and Decision Trees to complex models such as Transformers and Generative Adversarial Networks (GANs). Similarly, we choose code examples of varying lengths, ranging from small, 20 lines of code up to extensive, 800 lines of code. This diversity allows us to assess the adaptability and performance of LLMs across a broad range of MLOps challenges.

\end{itemize}

\subsection{Code Adaptation Methodology}

In this section, we describe our methodology to use LLMs for accomplishing the code \textit{Inlining} and \textit{Translation} tasks. The Inlining task involves prompt tuning along with temperature sampling and a DocPrompting method for difficult MLOps tasks. The Translation task involves a two-step process - a Data Curation Pipeline for relevant documentation extraction and a Prompt Construction Pipeline that enhances code translation using our proposed search approaches.

\begin{figure*}[t]
    \centering
    \includegraphics[width=\textwidth]{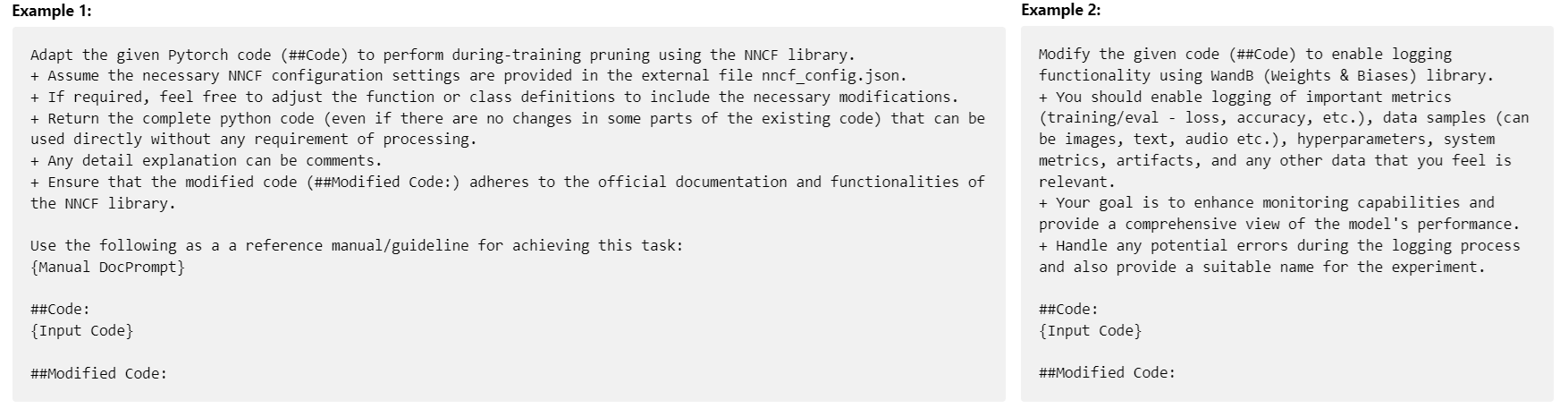} 
    \caption{Illustration of the Prompt Tuning Process for Model Optimization (Example 1) and Experiment Tracking (Example 2) Tasks. The $+$ sign shows the iterative process of adapting our prompt to identify the most effective prompt for accomplishing the intended MLOps task.}
    \label{fig:prompting}
    \end{figure*}
\subsubsection{Inlining}
For \textit{Inlining} tasks, we begin by conducting manual prompt tuning for each specific MLOps functionality under investigation. This involves a meticulous and iterative process where we experiment with different prompt variations and strategies to identify the most effective prompt for accomplishing the intended MLOps task. Fig. \ref{fig:prompting} demonstrates example prompt tuning for model optimization and experiment tracking tasks. To further assess the LLMs' capabilities, we use temperature sampling that modulates the randomness of the model's output. We experiment with three different temperature settings to gauge how they affect the performance of the LLMs in each of the code \textit{Inlining} MLOps tasks. In complicated MLOps tasks that demand a deep understanding of the MLOps component, we employ manual DocPrompting. This method involves using a comprehensive help document within the prompt, offering a step-by-step guide for accomplishing the specific MLOps task at hand. In our experimental results, we illustrate how incorporating such guidance documents can contribute to performance enhancement.

\subsubsection{Translation}
For achieving a specific MLOps functionality, the translation task involves translating code written  using one component, say X, to another component, say Y. For example, consider the following task for the LLM: to translate example code that involves version control using the GitPython library into equivalent code tailored for executing the same version control operations using the DVC Python library.

For accomplishing such complicated translation tasks, we design a DocPrompting based approach that involves a data curation and a prompt construction pipeline.
\begin{figure}[t]
\centering
\includegraphics[width=0.95\columnwidth]{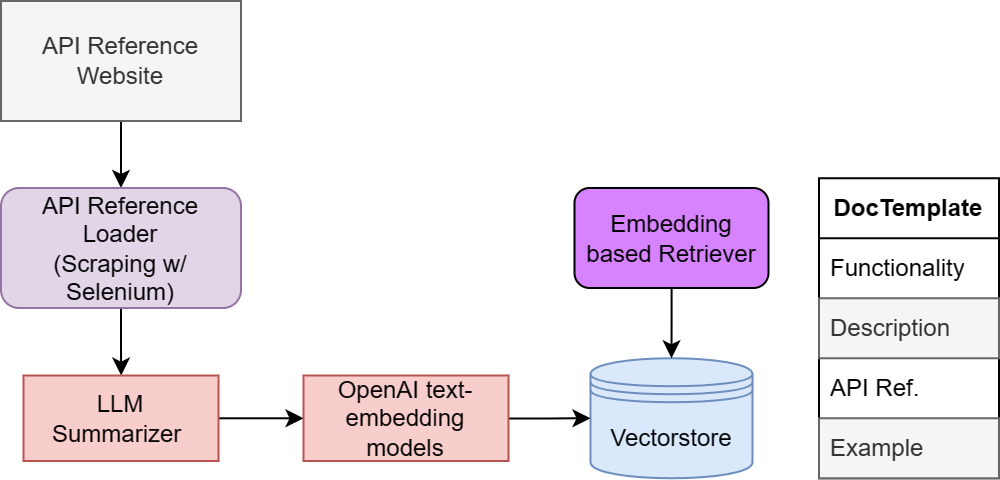} 
\caption{Translation Task - Data Curation Pipeline}
\label{fig1}
\end{figure}

\begin{itemize}
    \item \textbf{Data Curation Pipeline}: We use the Selenium automation framework to extract comprehensive documentation from both the source and target components. Initially, we use Selenium to scrape their documentation webpages with a fixed depth. Subsequently, we use a chunking process, segmenting these extensive documents into smaller, manageable sections. Since these crawled API documents can contain messy passages, we implement a post-processing step. We use an LLM to effectively summarize and condense the content of each of the chunks. The LLM's role is to distill the information into a concise passage that retains only the details relevant to functionalities of the component. This can include a functional description, API reference, and a usage example. These extracted and filtered chunks are then embedded into vector stores using OpenAI's Text-to-Embedding models. These vector stores are further used in the prompt construction pipeline for retrieving task-relevant documents.\\

    \item \textbf{Prompt Construction Pipeline}: For translating a X-Component based code (say, \textit{Code-X}) to a Y-Component based code, we hypothesize that having relevant documents from both X and Y components' documentation in the prompt can significantly enhance the translation task. To evaluate this hypothesis, we devise three approaches namely, DocsSearch, LLMSearch and LLM-DocsSearch, for retrieving the most relevant documents from X and Y's documentation vectorstores. In all three approaches, we initiate the process by conducting a top-k search of \textit{Code-X} within the vector store of Component-X to retrieve (\textit{Docs-X}), the most relevant documentation from Component-X.
    \label{sec:approach}
    \begin{itemize}
        \item \textbf{DocsSearch}: In this approach, we use the retrieved \textit{Docs-X} to retrieve relevant documentation from Component-Y (\textit{Docs-Y}) using a similar top-k search in Y's vectorstore. See Fig. \ref{fig:docssearch} for a visual explanation of this approach.
    \begin{figure*}[t]
    \centering
    \includegraphics[width=0.7\textwidth]{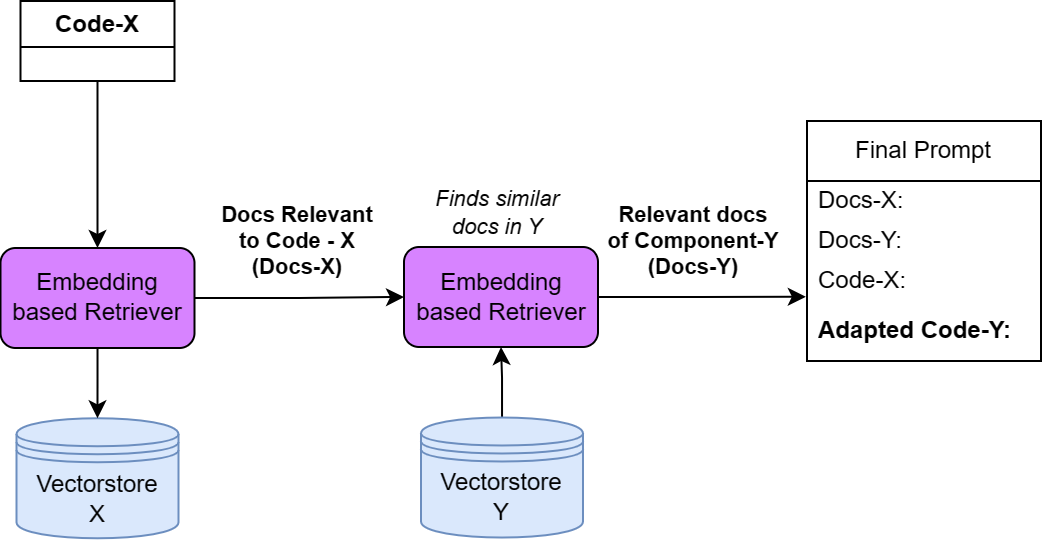} 
    \caption{DocsSearch Approach for Prompt Construction: This figure illustrates the DocsSearch approach within the Prompt Construction Pipeline. The approach leverages retrieved documentation (Docs-X) from Component-X to conduct a top-k search in Component-Y's vector store.}
    \label{fig:docssearch}
    \end{figure*}
        \item \textbf{LLMSearch}: In this approach, we forego the use of Component-Y's vector store for retrieving \textit{Docs-Y}. Instead, we directly experiment with an LLM to generate the relevant \textit{Docs-Y}, using the already retrieved \textit{Docs-X} as reference. Fig. \ref{fig:llmsearch} describes the LLMSearch approach.
    \begin{figure*}[t]
    \centering
    \includegraphics[width=0.7\textwidth]{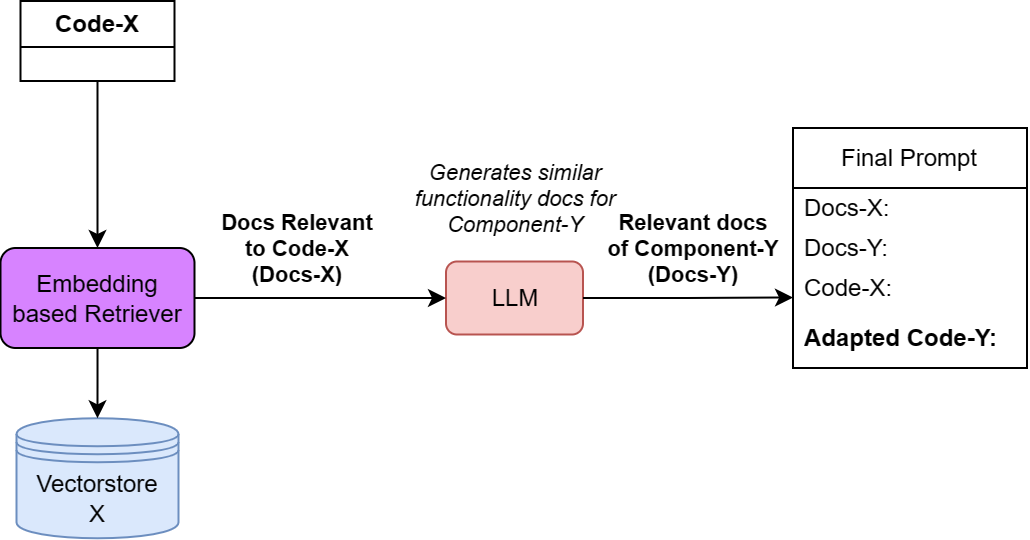} 
    \caption{LLMSearch Approach for Prompt Construction: This figure provides an overview of the LLMSearch approach within the Prompt Construction Pipeline. Unlike the DocsSearch method, LLMSearch bypasses the use of Component-Y's vector store to retrieve Docs-Y. Instead, it directly engages with a Large Language Model (LLM) to generate the relevant Docs-Y, using the previously retrieved Docs-X as a reference.}
    \label{fig:llmsearch}
    \end{figure*}
        \item \textbf{LLM-DocsSearch}: This approach combines both LLMSearch and DocsSearch techniques. Initially, we use an LLM to identify functionalities in Component-Y that bear similarity to those present in \textit{Docs-X}. Subsequently, the LLM-generated document is used to conduct a top-k search within the vectorstore of Component-Y, leading to the retrieval of the most relevant and accurate \textit{Docs-Y}. Fig. \ref{fig:llmdocssearch} shows the flow for this approach.
    \end{itemize}
\begin{figure*}[t]
\centering
\includegraphics[width=0.9\textwidth]{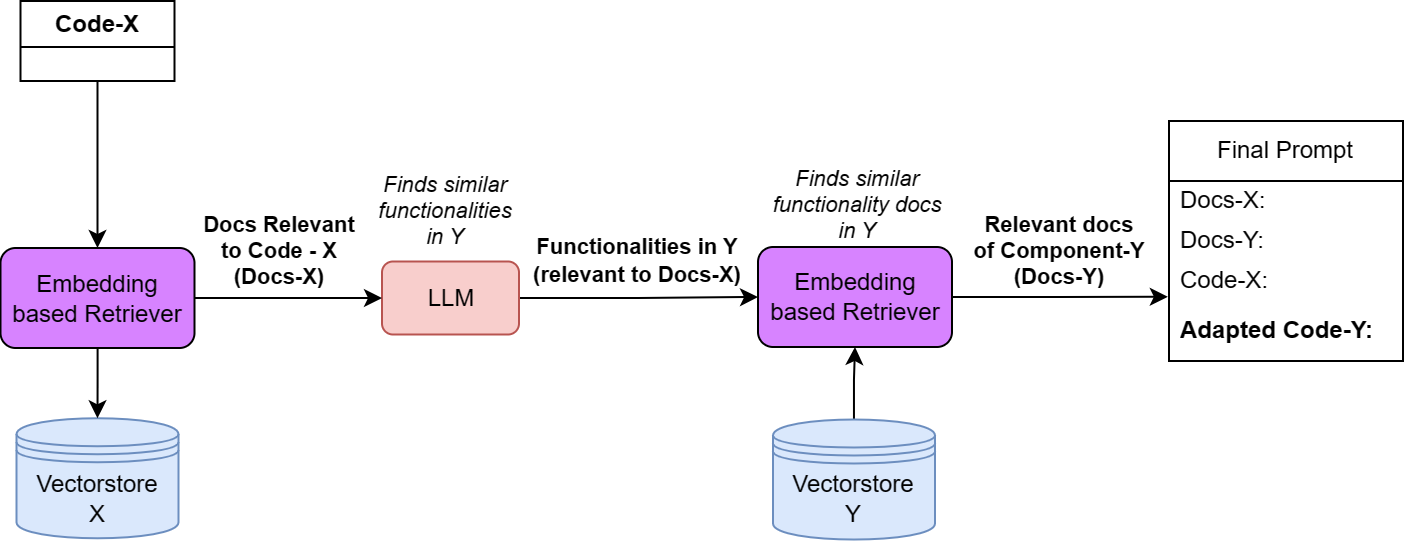} 
\caption{LLM-DocsSearch Approach for Prompt Construction: This figure illustrates the LLM-DocsSearch approach, a synthesis of the LLMSearch and DocsSearch techniques. Initially, the approach utilizes a Large Language Model (LLM) to identify similarities between functionalities in Component-Y and those present in Docs-X. Subsequently, the LLM-generated document is used to perform a top-k search within the vectorstore of Component-Y, facilitating the retrieval of the most relevant and accurate Docs-Y.}
\label{fig:llmdocssearch}
\end{figure*}
\end{itemize}

\subsection{Evaluation Method}
We use the Pass@3 evaluation metric, which implies that for each task with a temperature sampling value greater than 0, we consider three generations.  We systematically run all the generated code samples to verify their ability to execute the required task successfully. We consider a task successfully accomplished if at least one of these three generations results in accurate and executable code with minimal post-processing. Minimal post-processing includes instances where the LLM, having recognized parts of the code unaffected by task completion, skips them. In such cases, the LLM will generate only the necessary modifications, avoiding redundancy.


\section{Experiments and Results}

In this section, we describe our experiments and results of the benchmarking study comparing the performance of Open AI's gpt-3.5-turbo model with WizardCoder on the MLOps related code \textit{Inlining} and \textit{Translation} tasks.  To ensure the robustness of our experiments, we select widely-used task-specific components such as MLFlow and Weight \& Biases for experiment tracking, Optuna for hyperparameter optimization, etc.

\subsection{Inlining}
\subsubsection{Model Optimization}
\begin{figure}[t]
\centering
\includegraphics[width=0.99\columnwidth]{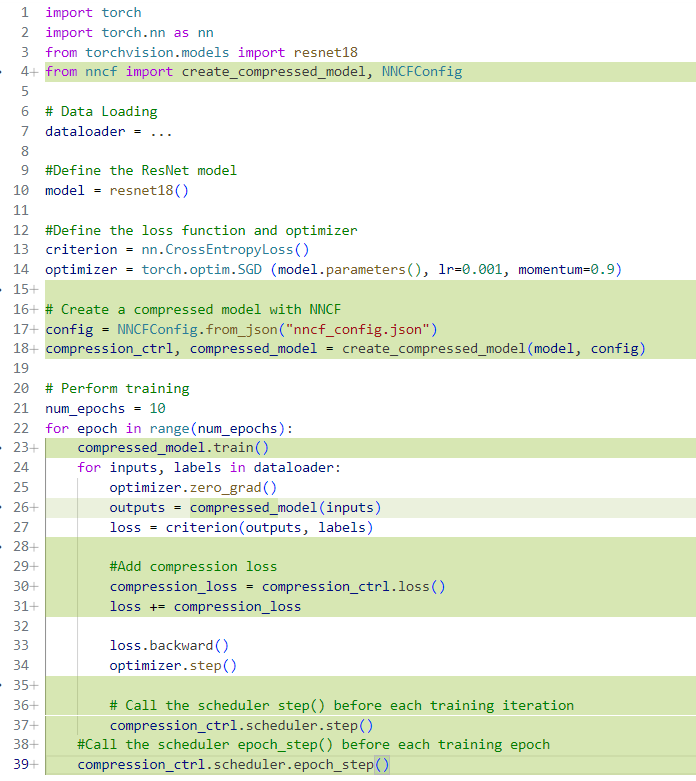} 
\caption{Code Inlining Task - The highlighted green sections demonstrate the expected inline adaptations when a simple model training script is provided as an input to an LLM.}
\label{fig:mo_example}
\end{figure}

In the context of this MLOps functionality, we focus on the during training pruning as the optimization task. Specifically, the objective is to enable Large Language Models (LLMs) to adapt existing code examples in-line to facilitate during training pruning. We test the adaptation ability using two distinct components: 1) Neural Network Compression Framework (NNCF) Library, 2) PyTorch Native functionalities.
Fig. \ref{fig:mo_example} shows the expected adaptions on a simple input example to incorporate during training pruning. To enhance the effectiveness of this task, we incorporate the Manual DocPrompting technique, as described in our Methodology section. This approach involves including relevant reference documentation in the prompt to guide the LLM in producing accurate code adaptations. Table \ref{tab:MO} shows a comparative analysis of the performance of OpenAI and WizardCoder models under various experimental settings. Each entry represents the percentage of the code examples that were accurately adapted and accomplish our desired task. This comparison allows us to assess the efficacy of different experimental settings while accomplishing this code \textit{Inlining} task.

\begin{table*}[ht]
\caption{Pass @3 Performance Comparison of OpenAI and WizardCoder Models in Code Inlining Task for During Training Pruning - Model Optimization. Each entry represents the percentage of code examples that were accurately adapted to accomplish the desired task.}
\centering
\begin{tabular}{l|lrrrrrr}
    \hline
 & \textbf{Temperature $\rightarrow$} & \multicolumn{2}{c}{\textbf{0}} & \multicolumn{2}{c}{\textbf{0.2}} & \multicolumn{2}{c}{\textbf{1}} \\ \hline
\textbf{Component} &  & \multicolumn{1}{c}{\textbf{Zero-Shot}} & \multicolumn{1}{c}{\textbf{DocPrompt}} & \multicolumn{1}{c}{\textbf{Zero-Shot}} & \multicolumn{1}{c}{\textbf{DocPrompt}} & \multicolumn{1}{c}{\textbf{Zero-Shot}} & \multicolumn{1}{c}{\textbf{DocPrompt}} \\ \hline
\multirow{2}{*}{NNCF} & WizardCoder & 0 & 0 & 0 & 0 & 0 & 0 \\
 & OpenAI & 0 & 0 & 0 & \textbf{50} & 0 & 30 \\ \hline 
\multirow{2}{*}{PyTorch-native} & WizardCoder & 0 & 0 & 0 & 0 & 0 & 10 \\
 & OpenAI & 20 & 40 & 50 & \textbf{60} & 50 & 60 \\ \hline
\end{tabular}
\label{tab:MO}
\end{table*}
    
\subsubsection{Experiment tracking}
In this MLOps functionality, we assess the LLMs' ability in enabling logging functionalities using the following two distinct components: 1) MLflow and 2) Weight \& Biases. The primary objective for LLMs in this context is to adapt existing code examples to enhance monitoring capabilities and provide a comprehensive overview of a model's performance. For our experiments with MLflow, we use code examples from three of the popular training frameworks: PyTorch Lightning, Keras, and scikit-learn. These frameworks are chosen due to their native autologging support within the MLflow ecosystem. Conversely, for Weights \& Biases experiments, we use PyTorch-based code examples. Table \ref{tab:ET} outlines the LLM benchmarks on different temperature settings. It is impressive to see the LLMs' ability to adapt highly complex and long code with such ease. We observe that as we increase the sampling temperature, the generated code displays significantly improved Experiment Tracking. They were able to capture a wider range of information, including different data modalities, parameters, hyperparameters, and more. For example, in a complex task that involved training a Deep Convolutional Generative Adversarial Network (DCGAN) \cite{dcgan}, the adapted code was even able to log the generated fake images at a fixed interval.

\begin{table*}[ht]
\caption{Pass@3 Performance Evaluation of LLMs in Code Inlining task for Experiment Tracking using MLflow and Weights \& Biases. Each entry represents the percentage of code examples that were accurately adapted to accomplish the desired task.}
\centering
\begin{tabular}{l|llrrr}
\hline
\textbf{Component} & \multicolumn{1}{c}{\textbf{ML Training Framework}} & \textbf{Temperature $\rightarrow$} & \multicolumn{1}{c}{\textbf{0}} & \multicolumn{1}{c}{\textbf{0.2}} & \multicolumn{1}{c}{\textbf{1}} \\ \hline
\multirow{6}{*}{MLflow} & \multirow{2}{*}{PyTorch Lightning} & WizardCoder & 50 & 75 & 75 \\
 &  & OpenAI & 100 & 100 & 100 \\
 & \multirow{2}{*}{Keras} & WizardCoder & 50 & 50 & 50 \\
 &  & OpenAI & 100 & 100 & 100 \\
 & \multirow{2}{*}{Sklearn} & WizardCoder & 50 & 25 & 75 \\
 &  & OpenAI & 75 & \textbf{100} & 75 \\ \hline
\multirow{2}{*}{Weights \& Biases} & \multirow{2}{*}{PyTorch} & WizardCoder & 10 & 30 & 50 \\
 &  & OpenAI & 100 & 100 & 100 \\ \hline
\end{tabular}
\label{tab:ET}
\end{table*}

\subsubsection{Model Registration}
In this MLOps functionality, we benchmark the ability of LLMs in adapting existing code examples based on training frameworks including PyTorch Lightning, Keras, and scikit-learn to perform model registration using MLflow. This enables tracking of different versions of the trained ML models, facilitating their retrieval and deployment as per specific requirements. Table \ref{tab:MR} shows the performance of both the LLMs in accomplishing this task across different temperature settings. We can see that the OpenAI model significantly outperforms the WizardCoder model. While tuning the prompt, we discovered that capitalization can be an effective strategy to emphasize specific aspects of the prompt. In this experiment, we observed that when a model is logged using MLflow, it is not registered in the MLflow Model Registry unless the 'registered-model-name' argument is provided. When we capitalized "UNLESS" while conveying this critical detail in the prompt, it consistently prompted the model to adhere to add this argument in a significant number of cases.

\begin{table*}[ht]
\caption{Pass@3 Performance Evaluation of LLMs in Code Inlining task for Model Registration using MLflow. Each entry represents the percentage of code examples that were accurately adapted to accomplish the desired task.}

\centering
\begin{tabular}{l|llrrr}
\hline
\textbf{Component} & \multicolumn{1}{c}{\textbf{ML Training Framework}} & \textbf{Temperature $\rightarrow$} & \multicolumn{1}{c}{\textbf{0}} & \multicolumn{1}{c}{\textbf{0.2}} & \multicolumn{1}{c}{\textbf{1}} \\ \hline
\multirow{6}{*}{MLflow} & \multirow{2}{*}{PyTorch Lightning} & WizardCoder & 0 & 0 & 50 \\
 &  & OpenAI & 75 & 50 & 75 \\
 & \multirow{2}{*}{Keras} & WizardCoder & 0 & 25 & 25 \\
 &  & OpenAI & 50 & 75 & 100 \\
 & \multirow{2}{*}{Sklearn} & WizardCoder & 0 & 25 & 50 \\
 &  & OpenAI & 75 & 100 & 100 \\ \hline
\end{tabular}
\label{tab:MR}
\end{table*}

\subsubsection{Hyperparameter Optimization}
In this MLOps functionality, we aim to benchmark the ability of LLMs to incorporate Hyperparameter Optimization into existing code examples. In our experiments, we prompt the LLMs to accomplish this functionality using the Optuna library, which is an automatic hyperparameter optimization framework. The adapted code thereby should use Optuna functionalities to search for the optimal hyperparameter configuration and then proceeds with the training of models using these optimal settings.

We use code examples based on PyTorch Lightning, Keras, and scikit-learn training frameworks. Table \ref{tab:HO} shows the performance of both the LLMs in accomplishing this task across different temperature settings. While performing Hyperparamter Optimization, the LLM has to significantly change the structure of the code, because it is first supposed to find the optimal hyperparameters by performing a search using an objective function and then train the model on the optimal hyperparameters. It is impressive to see how the gpt-3.5-turbo model is able to comprehend this information and accurately accomplish this MLOps fucntionality in at least 75\% of the examples (when temperature is 1).

\begin{table*}[ht]
\caption{Pass@3 Performance Evaluation of LLMs in Code Inlining task for Hyperparamter Optimization using Optuna. Each entry represents the percentage of code examples that were accurately adapted to accomplish the desired task.}
\centering
\begin{tabular}{l|llrrr}
\hline
\textbf{Component} & \multicolumn{1}{c}{\textbf{ML Training Framework}} & \textbf{Temperature $\rightarrow$} & \multicolumn{1}{c}{\textbf{0}} & \multicolumn{1}{c}{\textbf{0.2}} & \multicolumn{1}{c}{\textbf{1}} \\ \hline
\multirow{6}{*}{Optuna} & \multirow{2}{*}{PyTorch Lightning} & WizardCoder & 50 & 50 & 50 \\
 &  & OpenAI & 75 & 75 & 75 \\
 & \multirow{2}{*}{Keras} & WizardCoder & 25 & 50 & 50 \\
 &  & OpenAI & 75 & 75 & 75 \\
 & \multirow{2}{*}{Sklearn} & WizardCoder & 75 & 50 & 75 \\
 &  & OpenAI & 75 & 100 & 100 \\ \hline
\end{tabular}
\label{tab:HO}
\end{table*}

Through this assessment, we acquire valuable insights into the potential of LLMs in automating several critical MLOps functionalities. By accomplishing automation in these tasks that traditionally require manual intervention and time-consuming efforts, LLMs accelerate the entire Machine Learning workflow. This increased efficiency can lead to more rapid model development, testing, and deployment.

\subsection{Translation}
\subsubsection{Version Control}
\begin{figure}[t]
\centering
\includegraphics[width=0.99\columnwidth]{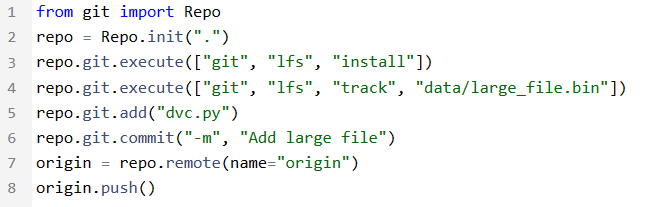} 
\caption{Example Version Control code written using the GitPython library.}
\label{fig:codeX}
\end{figure}

\begin{figure}[t]
\centering
\includegraphics[width=0.99\columnwidth]{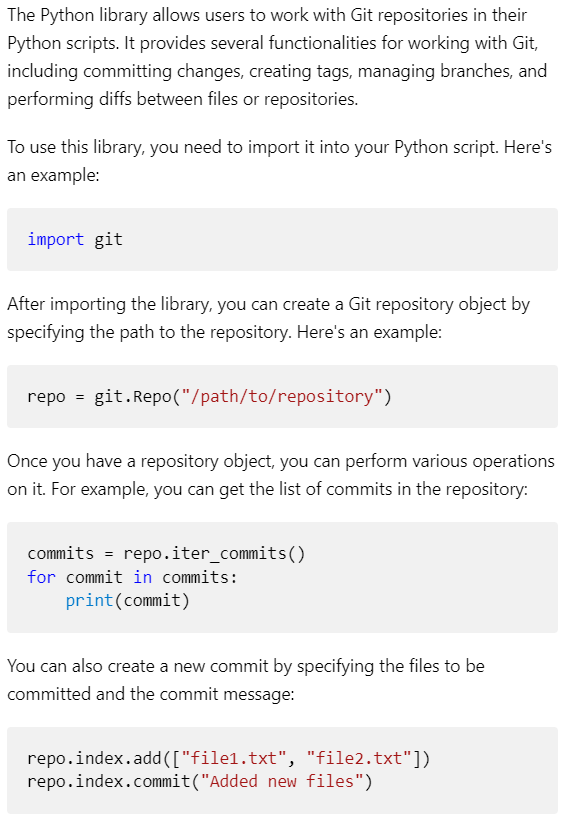} 
\caption{Relevant documents for \textit{Code-X} (Fig. \ref{fig:codeX}), retrieved from the vectorstore (\textit{Docs-X}) of GitPython library generated using our Data Curation Pipeline.}
\label{fig:docsX}
\end{figure}

\begin{figure}[t]
\centering
\includegraphics[width=0.99\columnwidth]{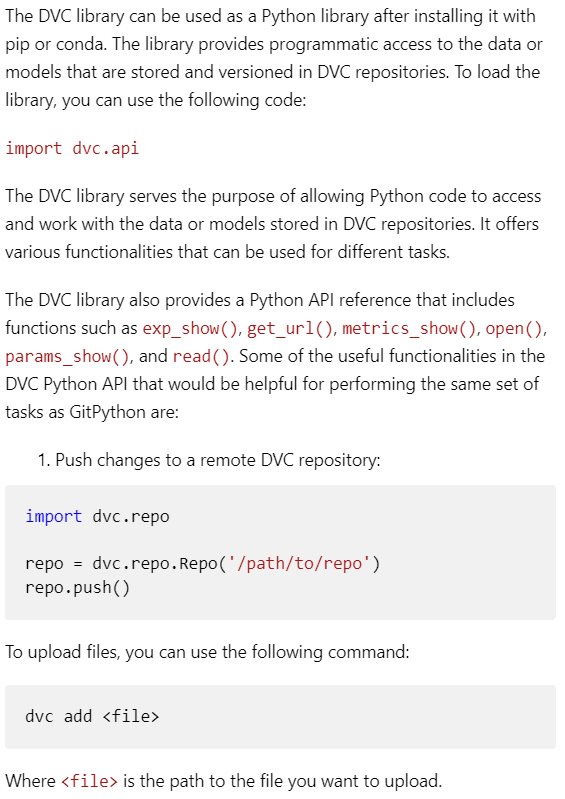} 
\caption{Relevant documents of DVC API for translating \textit{Code-X} (Fig. \ref{fig:codeX} ) - to perform data version control using DVC API, retrieved using our LLM-DocsSearch approach.}
\label{fig:docsY}
\end{figure}

In this section, we present our outcomes for the \textit{Translation} task related to Version Control MLOps functionality. The primary objective of LLMs in this context is to translate a given code example from one version control component to another, specifically, code based on the GitPython library into equivalent code suited for executing the same version control operations using the DVC Python library. 
Fig. \ref{fig:codeX} shows an example code (\textit{Code-X}) that uses the GitPython library to perform basic version control. In the data curation pipeline, we use the corresponding reference documentation of GitPython\footnote{https://gitpython.readthedocs.io/en/stable/} and DVC API\footnote{https://dvc.org/doc/}. For all three of our approaches, described in the Methodology section, we first retrieve \textit{Docs-X}, as shown in Fig. \ref{fig:docsX}. For effective translation, the constructed prompt (especially, \textit{Docs-Y} - the retrieved target component documentation) should contain highly relevant help snippets. Fig. \ref{fig:docsY} shows an example document in \textit{Docs-Y}, constructed using LLM-DocsSearch approach. In our experiments, we observe that when using the DocsSearch approach, often the retrieved documents are not directly relevant to our translation task. On the other hand, the LLMSearch approach tends to retrieve documents with inaccurate information due to the hallucinatory nature of LLMs. However, with the LLM-DocsSearch approach, we present a more effective method to retrieve pertinent information from the documentation of the target component. This approach intelligently harnesses the power of LLMs to enhance the retrieval process, resulting in the procurement of accurate and relevant snippets. When integrated into our prompts, these snippets significantly enhance the success of our \textit{Translation} task.

\section{Related Work}
Our work draws inspiration from a diverse set of program synthesis works. \cite{humaneval} introduced Codex, a GPT language model fine-tuned on publicly available code from GitHub, and examined its capabilities on a variety of coding
tasks. They also introduced HumanEval, an evaluation framework to measure functional correctness for synthesizing programs from docstrings. \cite{progsynth} explores the capabilities of LLMs in synthesizing small Python programs from natural language descriptions by introducing two benchmarks, Mostly Basic
Programming Problems (MBPP) and MathQA-Python. \cite{univmath} showcases an application of LLMs to solve, explain and generate university-level mathematics questions demonstrating the effectiveness of Program Synthesis using few-shot learning. In contrast to the above works, we introduce two new task categories that assess the ability of LLMs to solve much more complex code-adaptation tasks in the context of MLOps.

More recently, a significantly high number of LLMs have been released for code-related tasks, both closed-source - OpenAI Code-Davinci, Google PaLM-Coder \cite{palm} and open-source - BigCode StarCoder \cite{starcoder}, Microsoft WizardCoder \cite{luo2023wizardcoder}, Meta LLaMa \cite{touvron2023llama}, Salesforce CodeT5+ \cite{wang2023codet5}, etc. In contrast to the evaluations of these models on widely recognized benchmarks such as HumanEval, MBPP, etc., in this work, we perform a benchmarking study on complex code adaptation tasks involving large length MLOps based code examples.


\section{Future Work}

While this paper sheds light on the capabilities and limitations of existing LLMs for incorporating MLOps functionalities, some promising directions for future research in this domain can be fine-tuning open-source models to improve their performance. 

The prompt construction pipeline can be useful in generating custom datasets for any specific MLOps component. The extensible nature of the proposed \textit{Code Translation} techniques open up possibilities for seamlessly reconfiguring existing MLOps pipelines to incorporate alternative components offering equivalent MLOps functionalities. For instance, one can easily adapt Weight \& Biases based Experiment Tracking code to utilize alternative solutions like CometML or MLflow. Similarly, a tensorflow-based training job deployment configuration files, that define the essential resources for managing and executing an ML training job, can be easily adapted to PyTorch-based equivalents. As another example, one might utilize this methodology to adapt the Kubernetes manifests for deploying ML training jobs in a certain framework such as Kubeflow Job Operator to another framework such as run.ai. 


In this work, we only use automated evaluation for verifying the syntactic correctness of generated code. Future work may involve developing a comprehensive benchmarking framework, facilitating a systematic comparison of the emerging Code LLMs. Such a framework could aid in evaluating the effectiveness of these models across a wide range of MLOps scenarios. As the field of MLOps continues to evolve, leveraging LLMs for automating various tasks falling within the purview of code \textit{Inlining} and \textit{Translation}, will become extremely useful.

\section{Conclusion}
This article introduces the application of LLMs as a tool for enhancing critical MLOps functionalities. Using a benchmarking study, we show the potential of current state-of-the-art LLMs in enabling ML Engineers to adapt their existing codebases to diverse MLOps functionalities (\textit{Code Inlining}). Furthermore, we demonstrate methods to equip LLMs with the ability to comprehend complex API documentation, which opens up several avenues for real-world LLM applications. We envision our proposed methodologies for the \textit{Code Translation} task, including DocsSearch, LLMSearch, and LLM-DocsSearch, to significantly aid organizations in the seamless adaptation of code originally designed for one specific MLOps component, whether proprietary or open-source, to be compatible with alternative components. Particularly, considering the limitation posed by LLMs lacking prior knowledge of target components, our approach offers a promising method for effectively bridging this gap, enabling a smoother transition between different MLOps components.

In our study, we observe an inferior performance of WizardCoder, an open-source model, compared to the OpenAI gpt-3.5.turbo in complex code adaptation tasks. However, we anticipate the newer generations of LLMs such as LLama-2 (70B) \cite{llama2}, WizardCoder-Python (34B), Code Llama (34B) \cite{codellama} etc., to perform on par with the closed-sourced Open AI models.
In any case, we are certain that the methodologies that we follow for the code adaptation as well as for the benchmarking will be extremely valuable to the LLM research, ML Engineering, and DevOps communities.





\bibliographystyle{IEEEtran} 
\bibliography{ieeetai.bib}

\section*{Appendix}

\subsection{Dataset}
\subsubsection{PyTorch}
\begin{itemize}
    \item \url{https://github.com/pytorch/examples/tree/main/mnist}
    \item \url{https://github.com/pytorch/examples/tree/main/imagenet}
    \item \url{https://github.com/pytorch/examples/tree/main/dcgan}
    \item \url{https://github.com/pytorch/examples/tree/main/vision_transformer}
    \item \url{https://pytorch.org/tutorials/beginner/translation_transformer.html}
    \item \url{https://github.com/pytorch/examples/tree/main/vae}
    \item \url{https://github.com/pytorch/examples/tree/main/gcn}
    \item \url{https://github.com/pytorch/examples/tree/main/siamese_network}
    \item \url{https://github.com/pytorch/examples/tree/main/fast_neural_style/neural_style}
    \item \url{https://github.com/pytorch/examples/tree/main/word_language_model}
\end{itemize}
\subsubsection{PyTorch Lightning}
\begin{itemize}
\item \url{https://github.com/yang-zhang/lightning-language-modeling/tree/main}
\item \url{https://github.com/Lightning-AI/lightning/blob/master/examples/pytorch/domain_templates/reinforce_learn_Qnet.py}
\item \url{https://github.com/Lightning-AI/lightning/blob/master/examples/pytorch/domain_templates/generative_adversarial_net.py}
\item \url{https://github.com/Lightning-AI/lightning/blob/master/examples/pytorch/domain_templates/imagenet.py}
\end{itemize}

\subsubsection{Keras}
\begin{itemize}
\item \url{https://keras.io/examples/audio/transformer_asr/}
\item \url{https://keras.io/examples/rl/deep_q_network_breakout/}
\item \url{https://keras.io/examples/vision/mnist_convnet/}
\item \url{https://keras.io/examples/vision/super_resolution_sub_pixel/}
\end{itemize}

\subsubsection{sklearn}
\begin{itemize}
\item \url{https://scikit-learn.org/stable/auto_examples/tree/plot_tree_regression.html#sphx-glr-auto-examples-tree-plot-tree-regression-py}
\item \url{https://scikit-learn.org/stable/auto_examples/applications/plot_digits_denoising.html#sphx-glr-auto-examples-applications-plot-digits-denoising-py}
\item \url{https://scikit-learn.org/stable/auto_examples/linear_model/plot_sgd_early_stopping.html#sphx-glr-auto-examples-linear-model-plot-sgd-early-stopping-py}
\item \url{https://scikit-learn.org/stable/auto_examples/gaussian_process/plot_gpr_co2.html#sphx-glr-auto-examples-gaussian-process-plot-gpr-co2-py}
\end{itemize}

\end{document}